\documentclass[letter, 10 pt, conference]{ieeeconf}  

\IEEEoverridecommandlockouts                   

\usepackage{amssymb}
\usepackage{mathtools}
\usepackage{graphicx}
\usepackage{float}
\usepackage{array}
\usepackage{tikz}
\usepackage{listings}
\usepackage{hyperref}
\usepackage{graphicx}
\usepackage{cite}
\usepackage{dsfont}
\usepackage{mathtools,etoolbox}
\usepackage[ruled, linesnumbered]{algorithm2e}
\usepackage[none]{hyphenat}
\usepackage[utf8]{inputenc}
\usepackage[english]{babel}
\usepackage[T1]{fontenc}
\usepackage{mathtools, nccmath}

\usepackage{amsthm}
\usepackage{xfrac}
\usepackage{nicefrac}
\usepackage{booktabs}
\usepackage[switch, pagewise]{lineno}
\usepackage{mathrsfs}
\DeclareMathAlphabet{\mathpzc}{OT1}{pzc}{m}{it}
\usepackage{multirow}
\usepackage{multicol}
\usepackage{bbm}
\usepackage{soul}
\usepackage{xfrac}
\usepackage{balance}
\usepackage{titlesec}
\usepackage[final]{pdfpages}
\usepackage{amsmath}
\usepackage{xcolor}

\DeclareMathOperator*{\argmax}{argmax} 

\definecolor{ultramarine}{RGB}{68,114,196}

\hypersetup{colorlinks=true,urlcolor=blue,citecolor=red}
\usepackage[some, placement=top]{background}

\backgroundsetup{
        placement=top,
        position={8cm,2cm}}

\SetBgScale{1}
\SetBgContents{\parbox{16cm}{\Large \centering This paper has been accepted for publication at the IEEE International Conference on Robotics and Automation (ICRA),  Paris, 2020. \textcopyright IEEE}}
\SetBgColor{gray}
\SetBgAngle{0}
\SetBgOpacity{0.9}

\newcommand{\norm}[1]{\left\lVert#1\right\rVert} 

\title{\LARGE \bf NudgeSeg: Zero-Shot Object Segmentation\\ by Repeated Physical Interaction}

\author{Chahat Deep Singh*, Nitin J. Sanket*, Chethan M. Parameshwara, Cornelia Ferm{\"u}ller, Yiannis Aloimonos 
\thanks{All the authors are with the Perception and Robotics Group, University of Maryland Institute for Advanced Computer Studies, University of Maryland, College Park. *\textit{Equal Contribution}. (Corresponding author: Chahat Deep Singh)}} 

\begin{document}

\makeatletter
\g@addto@macro\@maketitle{
\begin{figure}[H]
  \setlength{\linewidth}{\textwidth}
  \setlength{\hsize}{\textwidth}
    \centering
    \includegraphics[width=\textwidth]{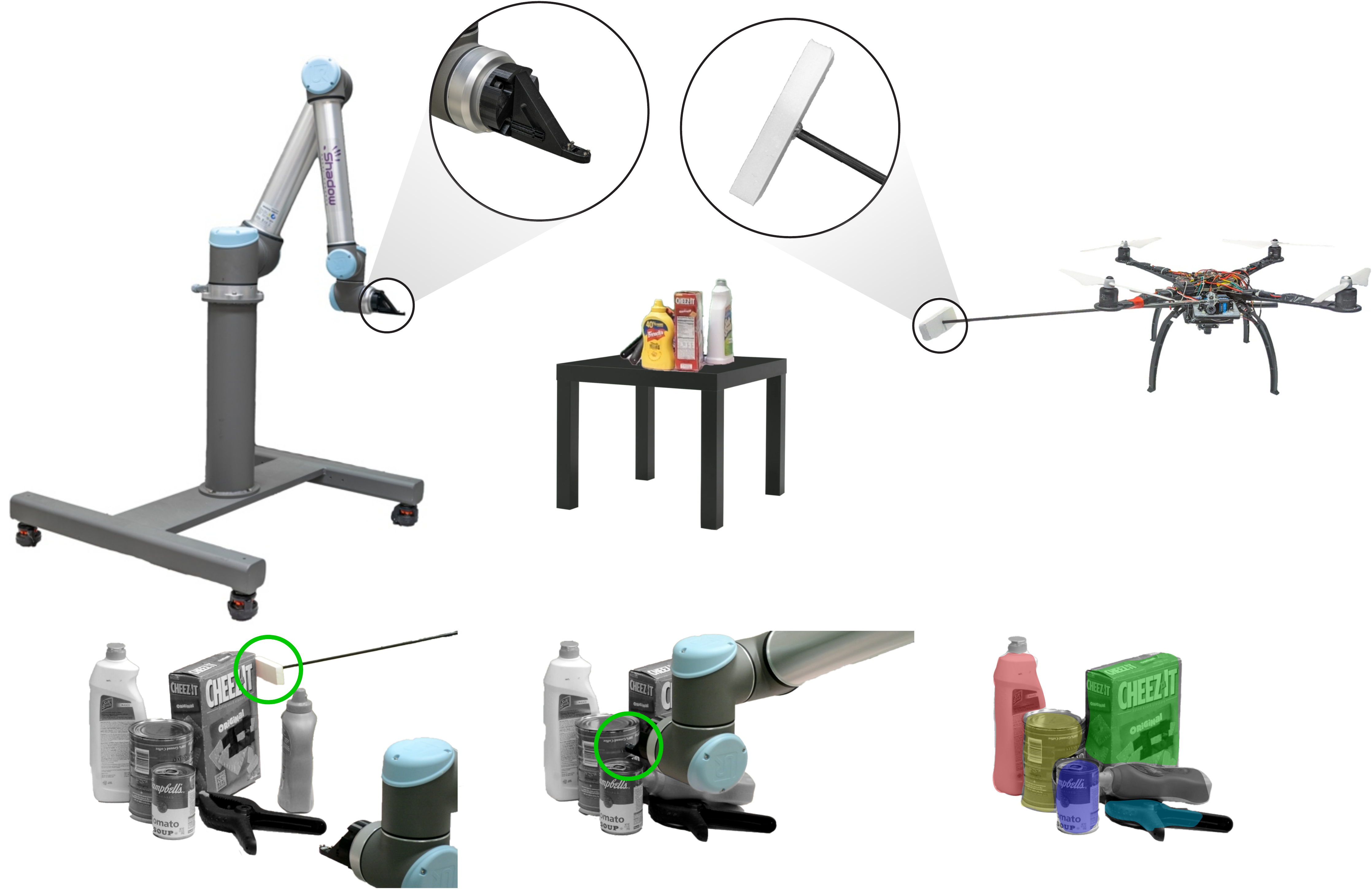}
    \caption{Top row: Robots (UR-10 and a quadrotor) used to physically interact (or nudge) with the objects to get motion cues for segmenting objects in a clutter. Bottom row (left to right): Initial Configuration of a cluttered scene and the first nudge being invoked, final nudge is invoked, final Segmentation of the cluttered scene. Green circles show the nudge operation. \textit{All the images in this paper are best viewed in color at 200\% zoom on a computer screen.}}
    \vspace{-20pt}
    \label{fig:Banner}
    \end{figure}
}
\maketitle
\thispagestyle{empty}
\pagestyle{empty}

\setcounter{figure}{1}

\begin{abstract}

Recent advances in object segmentation have demonstrated that deep neural networks excel at object segmentation for specific classes in color and depth images. However, their performance is dictated by the number of classes and objects used for training, thereby hindering generalization to never seen objects or \textit{zero-shot samples}. To exacerbate the problem further, object segmentation using image frames rely on recognition and pattern matching cues. Instead, we utilize the `active' nature of a robot and their ability to `interact' with the environment to induce additional geometric constraints for segmenting zero-shot samples.


In this paper, we present the first framework to segment unknown objects in a cluttered scene by repeatedly `nudging' at the objects and moving them to obtain additional motion cues at every step using only a monochrome monocular camera. We call our framework \textit{NudgeSeg}. These motion cues are used to refine the segmentation masks. We successfully test our approach to segment novel objects in various cluttered scenes and provide an extensive study with image and motion segmentation methods. We show an impressive average detection rate of over $86\%$ on zero-shot objects.

\end{abstract}
\section*{Supplementary Materials}
The supplementary video is available at \url{http://prg.cs.umd.edu/NudgeSeg}.

\section{Introduction and Philosophy}

\begin{figure}[t!]
    \centering
    \includegraphics[width=\columnwidth]{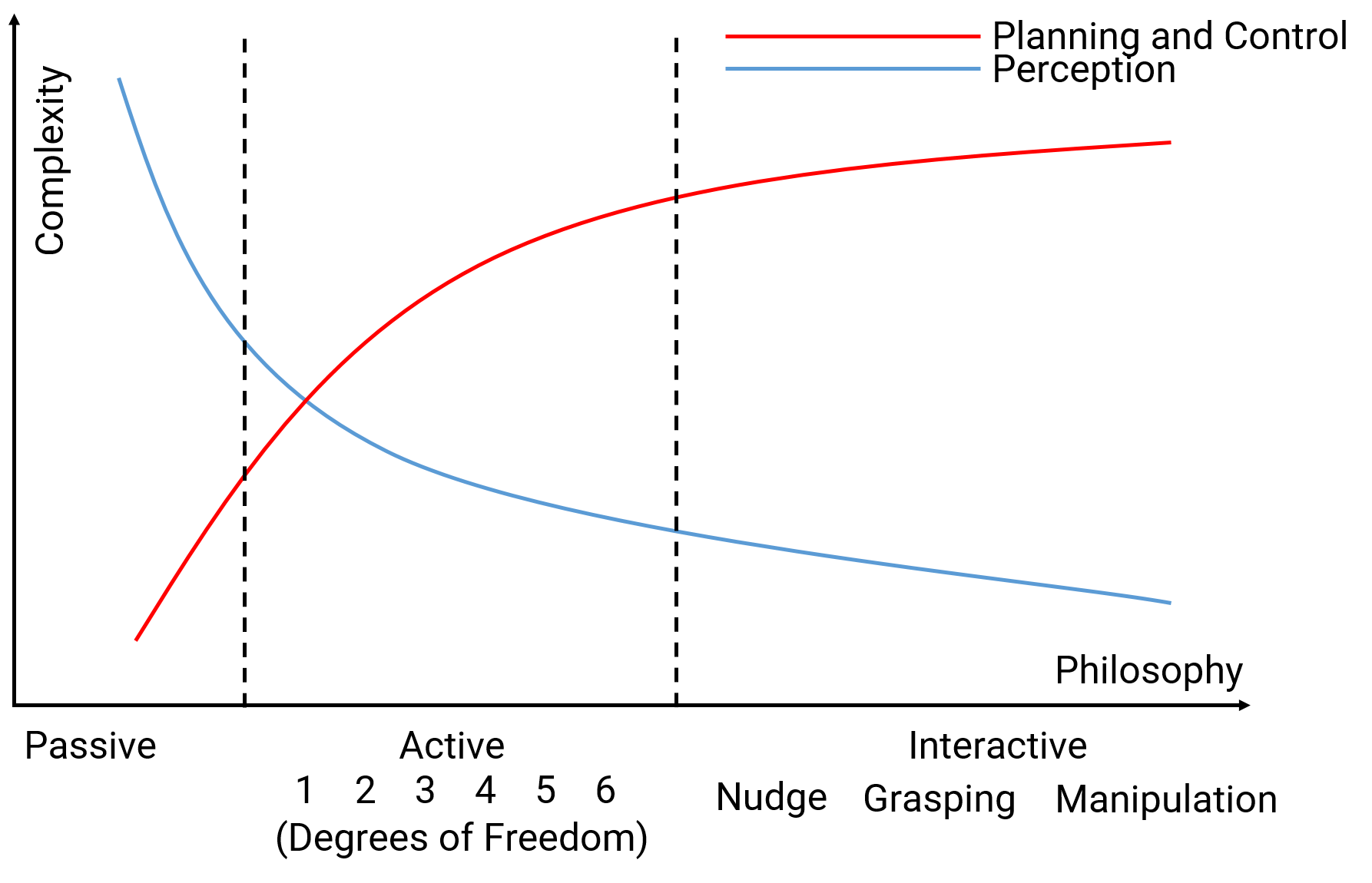}
    \caption{A conceptual graph of variation of complexity in perception, planning and control with task philosophy. As a keen observation, the algorithmic complexity decreases with increase in the manipulator motion.}
    \label{fig:PhilosophyGraph}
\end{figure}

\textit{Perception} and \textit{Interaction} form a complimentary synergy pair which is seldom utilized in robotics despite the fact that most robots can either move or move a part of their bodies to gather more information. This information when captured in a \textit{smart} way helps simplify the problem at hand which is expertly utilized by nature's creations. Even the simplest of biological organisms relies on this active-interactive synergy pair to simplify complex problems \cite{egelhaaf2012spatial}. To this end, the pioneers of robotics laid down the formal foundations that captured the elegance of action-interaction-perception loops. The amount of computation required for performing a certain task can be supplemented by exploration and/or interaction to obtain information in a manner that simplifies the perception problem. Fig. \ref{fig:PhilosophyGraph} shows a representative plot of how complexity in perception, planning and control vary with different design philosophies. We can observe that there are different amounts of activeness captured by the number of degrees of freedom in which the agent can move. The amount of interactiveness can vary from being able to nudge to grasp (pick-up) to complex manipulation (screwing a lid). One can trade off the complexity in planning and control to that of perception by the choice of task philosophy. 

We present a framework that captures this elegance to address the problem of segmentation of objects of unknown shape and type (also called \textit{zero-shot objects}). Such a method would serve as an initial guess to learn new objects on a robot by interacting with it -- just like how babies learn about new objects. To our knowledge, this is the first work which addresses the problem of zero-shot object segmentation from clutter by iterative interaction using a grayscale camera. We formally describe our problem statement with a list of key contributions next.








\subsection{Problem Formulation and Contributions}
The question we address is as follows: \textit{How can we segment objects of unknown shape and type (\textit{zero-shot samples}) from a cluttered scene by interacting (\textit{nudging}) with the objects using a monochrome monocular camera?} \\
The key contributions of this paper are given below:

\begin{itemize}
    \item We propose an active-interactive nudging framework called \textit{NudgeSeg} for segmenting zero-shot objects from clutter using a monochrome monocular camera.
    \item Conceptualization of uncertainty in optical flow to find the object clutter pile.
    \item Extensive real-world experiments on different robots including a quadrotor and a  robotic arm to show that our framework is agnostic to the robot's structure.  
\end{itemize}

We make the following explicit assumptions in our work:
\begin{itemize}
    \item  The surface on which the cluttered object pile is located is planar. 
    \item All the individual objects in the clutter are non-deformable.
\end{itemize}


\subsection{Related Work}

We subdivide the related work into three parts: instance and semantic segmentation, motion segmentation and active--interactive conceptualization. 

\textbf{Instance and Semantic Segmentation:} The last few years have seen a renewal of interest in instance and semantic segmentation after the deep learning boom. Long \textit{et al.} \cite{long2015fully} proposed the first approach to adopt fully Convolutional Neural Networks (CNNs) for semantic segmentation. Later, CNN meta-architecture approaches \cite{he2017mask} occupied top spots in recent object segmentation challenges. These were followed by TensorMask \cite{chen2019tensormask} that used an alternative sliding-window network design to predict sharp high-resolution segmentation masks. Recently, a region-based segmentation model was introduced in \cite{kirillov2020pointrend} that can produce masks with fine details pushing the accuracy of region-based approaches even further.

\textbf{Motion Segmentation:} The problem of segmenting motion into clusters which follow a similar 3D motion has been thoroughly studied in  \cite{torr1998geometric, tron2007benchmark}. Later, several methods were introduced that use an expectation maximization approach for segmentation \cite{517092, jepson1993mixture} to improve the results further for in the wild sequences. Some recent work \cite{bideau2018best, bideau2016s, wulff2017optical} address more complex scenarios with optical flow inputs, considering motion angle as the motion information for segmentation. Furthermore, there has been extensive progress in the field of event camera based motion segmentation in the past decade. These event camera-based approaches commonly utilize event alignment methods and expectation-maximization \cite{stoffregen2019event, 0-MMS} schemes to obtain segmentation masks for independently moving objects.

\textbf{Active and Interactive Approaches:}
The first conceptualization of Active vision was presented in Aloimonos \textit{et al.} \cite{ActiveVision} and Bajcsy \textit{et al.} \cite{BajcsyActive} where the key concept was to move the robot in an exploratory way that it can gather more information for the task at hand. The synergy pair to active approaches are the ones which involve interaction between the agent and its environment -- interactive approach which was formally defined in \cite{8007233}. For robots with minimal amount of computation, it's activeness' and `interactiveness' take precedence to gather more information in a way to simplify the perception problem. In other words, with more exploration information, one can trade-off the amount of sensors required or computational complexity to solve a given task. Such an approach has been demonstrated to clear or segment a pile of unknown objects using manipulators \cite{katz2013clearing, hausman2012segmentation, agrawal2016learning, TamimAsfour} utilizing depth and/or stereo cameras. Similar concepts of interaction have also been used to infer the object properties like shape and weight using only haptic feedback \cite{natale2004learning}. Recently, \cite{pathak2018learning} introduced the concept of learning to segment by grasping objects randomly in a self-supervised manner using color images. 

Our contribution draws inspiration from all the aforementioned domains and induces motion by iteratively nudging objects to generate motion cues to perform motion segmentation to segment never seen objects.

\subsection{Organization of the paper}
We describe our proposed \textit{NudgeSeg} segmentation framework in Sec. \ref{sec:NudgeSegFramework} in detail with subsections explaining each step of the process. In Sec. \ref{sec:Results} we present our evaluation methodology along with extensive comparisons with other state-of-the-art segmentation methods along with detailed analysis. We then provide discussion along with thoughts for future directions in Sec. \ref{sec:Discussion}. Finally, we conclude our work in Sec. \ref{sec:Conclusions}. 


\section{NudgeSeg Framework}
\label{sec:NudgeSegFramework}
Our framework utilizes both concepts of active and interactive perception for its different parts. The first part utilizes active perception ideology where the camera is moved to obtain a hypothesis of where the \textit{object pile} is located (foreground-background segmentation). Next we utilize the interactive perception ideology to \textit{repeatedly nudge/poke objects} to gather more information for obtaining a segmentation hypothesis. Both the parts are explained in detail next.



\subsection{Active perception in NudgeSeg}
\label{subsection:ActiveApproach}
As explained earlier, the precursor to object segmentation is to \textit{find the pile of objects}. Since, our robot(s) is not equipped with a depth sensor, there is no trivial way to obtain the segmentation of the object pile. Hence, we utilize the activeness of the robot  to obtain a function correlated with depth by moving the robot's camera. 





Let the image captured at time index $i$ be denoted as $\mathcal{I}_i$. The dense pixel association matrix also called optical flow \cite{horn1981determining} is given by $\dot{p}_{ij}$ between frames $i$ and $j$. Note that, these optical flow equations use the pinhole camera projection model.

The boundary between the foreground and background is correlated to the amount of occlusion a pixel ``experiences'' between two frames. This occlusion is inversely related to the condition number $\kappa$ of the optical flow estimation problem. Although there is no direct way to obtain $\kappa$, we can obtain the dual of this quantity, i.e., the estimated uncertainty $\rho$. In particular, we utilize  \textit{Heteroscadatic Aleatoric uncertainty} \cite{gast2018lightweight} since it can be obtained without much added computation overhead. Details of how $\rho$ is obtained from a network is explained in Sec. \ref{subsection:Network}.

Once, we obtain $\rho$ between two frames, the next step is to find the \textit{first nudge direction}. To accomplish this, we perform morphological operations on $\rho$ to obtain the blobs which are a subset of the object pile $\{\mathcal{O}_k\}$ ($k$ is the blob index). We then find the convex hull of this set $\mathcal{C}\left(\{\mathcal{O}\}\right)$. The first poke direction is obtained in two steps. First, we find the blob which has the highest average uncertainty value (since it will have the lowest noise in $\rho$) which is mathematically given as $ \argmax_k \mathbb{E}\left( \rho\left(\mathcal{O}_k\right)\right)$. We then compute the first nudge point $\mathpzc{N}_1$ which is given as the closest point to the centroid of blob $k$ we found earlier:

\begin{equation}
    \mathpzc{N}_1 = \argmax_{\mathbf{x}} \lVert \mathcal{C}(\mathcal{O}_\kappa)_{\mathbf{x}} - \overline{(\mathcal{O}_\kappa)} \rVert_2
\end{equation}

\begin{figure*}[t!]
    \centering
    \includegraphics[width=\textwidth]{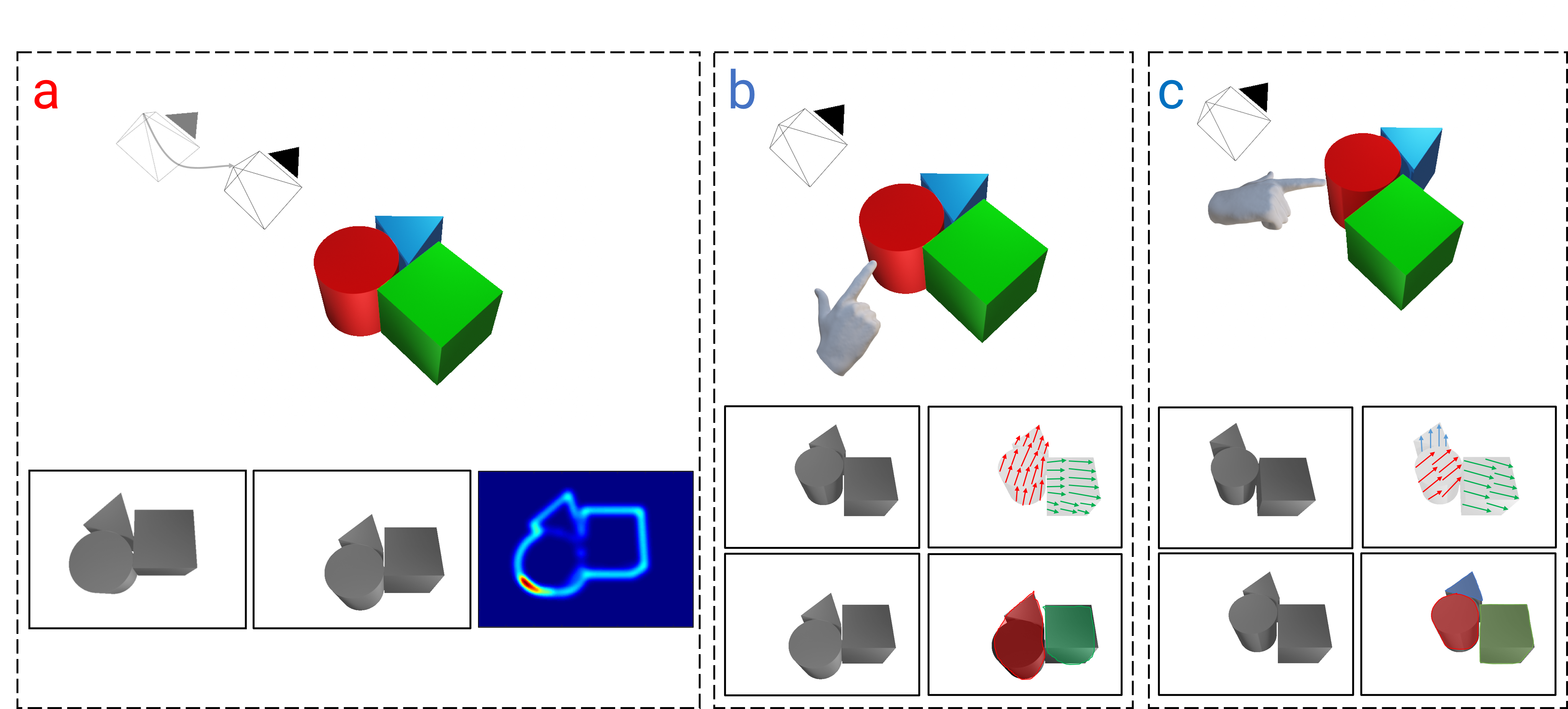}  
    \caption{{\color{red}(a)} \textit{Active} perception in \textit{NudgeSeg} framework. Top row shows the movement of the camera. Bottom row shows the image inputs and uncertainty $\rho$. {\textcolor{ultramarine}{(b)}} and {\textcolor{ultramarine}{(c)}} \textit{Interactive} perception in \textit{NudgeSeg} framework. Top row shows the object nudging. Bottom row shows the input images (before and after nudge), optical flow representation and segmentation hypothesis where colors indicate cluster membership.}
    \label{fig:Algo}
\end{figure*}

Here, $\mathbf{x}$ indexes each point in $\mathcal{C}\left(\mathcal{O}_k\right)$ and $\overline{A}$ denotes the centroid of $A$. We then nudge the objects at $\mathpzc{N}_1$ with our \textit{nudging tool} (See Fig. \ref{fig:Banner}). Fig. \ref{fig:Algo}{\color{red}(a)} summarizes the active segmentation part.

\begin{figure}
    \centering
    \includegraphics[width=0.7\columnwidth]{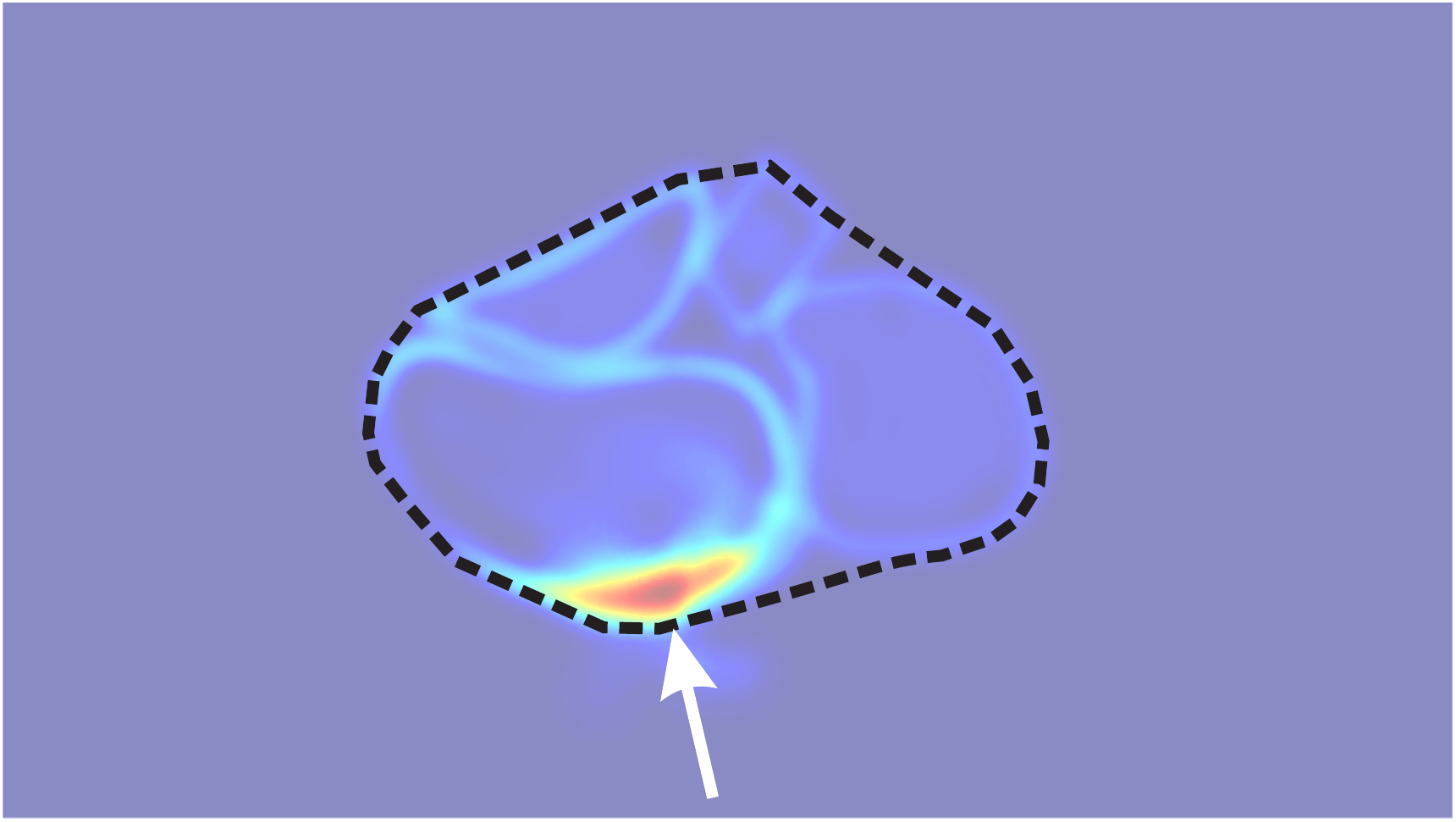}
    \caption{First nudge policy using uncertainty in optical flow. Hotter colors represents higher uncertainty. The dashed line represents the convex hull of the cluttered scene and the arrow represents the direction of first nudge at point $\mathpzc{N}_1$.}
    \label{fig:UncPoke}
\end{figure}

\subsection{Interactive perception in NudgeSeg}
\label{subsection:InteractiveSeg}

The basis of our interactive perception part is that we can cluster rigid parts of the scene using optical flow $\dot{p}$ (See Sec. \ref{subsection:Network} for details on how $\dot{p}$ is obtained). Since, we generally do not obtain a complete segmentation of the scene by clustering optical flow from a single nudge, we propose a iterative nudging strategy based on the statistics of the current cluster hypothesis (See Sec. \ref{subsubsection:NudgePolicy} for more details). 

In order to split the current scene based on optical flow (before and after first nudge) into multiple segments, we employ Density-based Spatial Clustering of Applications with Noise (DBSCAN) \cite{ester1996density} since it does not require the number of clusters as the input. The following criteria are used to assign two points $\mathbf{X}$ and $\mathbf{Y}$ to the same  cluster:

\begin{equation}
 \norm{\mathbf{X} - \mathbf{Y}}_2 < \tau_d 
\end{equation}

\begin{equation}
 \norm{\mathcal{M}_{\mathbf{X}} - \mathcal{M}_{\mathbf{Y}}}_2 < \tau_{\mathcal{M}}
\end{equation}

\begin{equation}
 \min\left({\lvert \mathcal{A}_{\mathbf{X}}} - \mathcal{A}_{\mathbf{Y}}\rvert, \,2\pi - {\lvert \mathcal{A}_{\mathbf{X}} - \mathcal{A}_{\mathbf{Y}}}\rvert\right) \leq \tau_{\mathcal{A}}
\end{equation}

Here, $\tau_{d}, \tau_{\mathcal{M}}$ and $\tau_{\mathcal{A}}$ are user-defined thresholds. The input to DBSCAN is 4-dimensional data consisting of the image coordinates, optical flow magnitude and direction: $\begin{bmatrix} \mathbf{x} & \mathcal{M} & \mathcal{A}\end{bmatrix}^T$. Points which do not belong to any cluster are removed as noise points since their density is very low. We repeat this step iteratively until we reach the termination criterion as described in Sec. \ref{subsection:Termination}. We denote segmentation hypothesis output at each iteration (also called time index $i$) as $\mathcal{H}_i$.\\

\subsubsection{Nudging Policy}
\label{subsubsection:NudgePolicy}
Firstly, we compute the covariance matrix $\{\Sigma_i^k\}$ of the optical flow $\dot{p}$ for all clusters of the hypothesis $\{\mathcal{H}_i^k\}$ ($k$ is the cluster index) and is given by

\begin{equation}
    \Sigma_i^k = \mathbb{E} \left(\left(\dot{p}_i^k - \mathbb{E}\left( \dot{p}_i^k\right)\right)\left(\dot{p}_i^k - \mathbb{E}\left( \dot{p}_i^k\right)\right)^T\right)
\end{equation}

We then use $\Sigma_i^k$ to find the Eigenvectors and Eigenvalues for each cluster given by $v_{\text{min}, i}^k, v_{\text{max}, i}^k$  and $\lambda_{\text{min}, i}^k, \lambda_{\text{max}, i}^k$. Then, we pick the cluster $k$ with the second largest condition number $\kappa_* = \nicefrac{\vert \lambda_{\text{max}}\vert}{\vert\lambda_{\text{min}} \vert}$ (since we do not want to nudge the object with the highest motion information, i.e., best $\kappa$). The nudging direction is chosen as the Eigenvector direction $v_{\text{min}}$ for the cluster chosen before. However, if $\kappa_* \le \tau$, we nudge the cluster with largest $\kappa$ since the quality of motion cue corresponding to $\kappa_*$ is noisy. Finally, we nudge using a simple Proportional-Integrative-Differential (PID) controller servoing the nudge point. After the current nudge $i$, we propagate the mask to the new frame $j$ by warping using optical flow. We call this propagated mask $\hat{\mathcal{H}}$.\\

\subsubsection{Mask Refinement}
\label{subsubsubsection:MaskRefinement}
Finally, for each nudge step, once the masks are propagated, the robot aligns itself to initiate the next nudge for new motion cues. After the next nudge is invoked, a new $\mathcal{H}$ is generated which may or may not overlap with the previous propagated masks. To find the updated masks, $\prescript{k}{}{}\mathcal{H}_j \cup \prescript{k}{}{}\mathcal{\hat{H}}_j$ is computed. Refer Algorithm \ref{alg:propagation} for the mask refinement details.  Fig. \ref{fig:Algo}{\color{red}(b)} and {\color{red}(c)} each represents one step (single nudge) of the \textit{interactive} segmentation part. 

\subsection{Verification and Termination}
\label{subsection:Termination}
If the mean IoU between $\mathcal{H}_{i}$ and $\mathcal{H}_{i+n}$ have not changed more than a threshold, we invoke the verification step. In this step, we nudge each segment independently in a direction towards its geometric center to ``see'' if the object splits further. This is performed once nudge for every cluster and if any segment splits into more than one, we go back to the procedure described in Sec. \ref{subsection:InteractiveSeg} for those clusters, if not, we terminate.

\begin{algorithm}[t!]
\caption{Segment Propagation And Mask Refinement}
\label{alg:propagation}
\SetAlgoLined
\KwData{Optical Flow $\dot{p}^{k}_{i}$, Segmentation Hypothesis of $k$ masks $\{\mathcal{H}_{i}\}$ in $i^{th}$ frame.}
\KwResult{Updated Segmentation Hypothesis $\{\mathcal{{H}}_{j}\}$ in $j^{th}$ frame.}
 \eIf{$\prescript{k}{}{}\mathcal{H}_j \cap \prescript{k}{}{}\mathcal{\hat{H}}_j > \tau_{\mathcal{H}}$}{
$\prescript{k}{}{}\mathcal{\hat{H}}_j \rightarrow \prescript{k+1}{}{}\mathcal{H}_j$ \hfill \CommentSty{Update Masks};
$\prescript{k}{}{}\mathcal{H}_j  \rightarrow \max(\prescript{k}{}{}\mathcal{\hat{H}}_j, \prescript{k}{}{}\mathcal{{H}}_j)  - (\prescript{k}{}{}\mathcal{\hat{H}}_j  \cap \prescript{k}{}{}\mathcal{{H}}_j)$\;
}{$\prescript{k}{}{}\mathcal{\hat{H}}_j \rightarrow \prescript{k+1}{}{}\mathcal{{H}}_j$ \hfill \CommentSty{Create New $\mathcal{H}$\;}}
\end{algorithm}


\subsection{Network Details}
\label{subsection:Network}
We obtain optical flow $\dot{p}$ using a Convolutional Neural Network (CNN) based on PWC-Net \cite{sun2018pwc}. We use the multi-scale ($L$ scales) training loss given below:

\begin{equation}
    \mathcal{D}\left(\hat{\dot{p}}_l, \tilde{\dot{p}}_l  \right) = \sum_{\forall l} \alpha_l \mathbb{E}_{\mathbf{x}}\left( \Vert {\hat{\dot{p}}}_l\left(\mathbf{x}\right) - {\tilde{\dot{p}}}_l\left(\mathbf{x}\right)\Vert_1\right) 
\end{equation}

where $l$ is the pyramid level, $\hat{\dot{p}}_l$ and $\tilde{\dot{p}}_l$ denotes the ground truth and predicted optical flows at level $l$ respectively. $\alpha_l$ is the weighing parameters for $l^{\text{th}}$ level. We refer the readers to \cite{sun2018pwc} for more network details. 

Finally, to obtain the \textit{Heteroscadatic Aleatoric uncertainty} \cite{gast2018lightweight}, we change the output channels from two to four in the network, i.e., $\tilde{\hat{p}}$ (two channels) and the predicted uncertainty $\rho$ (two channels, one for each channel $\tilde{\hat{p}}$) is obtained in a self-supervised way and is trained using the following final loss function

\begin{multline}
    \mathcal{L} = \sum_{\forall l} \alpha_l  \mathbb{E}_{\mathbf{x}} \left( \frac{ \Vert {\hat{\dot{p}}}_l\left(\mathbf{x}\right) - {\tilde{\dot{p}}}_l\left(\mathbf{x}\right)\Vert_1}{\log_e \left(1 + e^{\left(\rho_l\left( \mathbf{x}\right)+\epsilon\right)}\right)} \right)  +\\  \sum_{\forall l}\alpha_l\mathbb{E}_{\mathbf{x}}\left( \log_e\left(1 + e^{\rho_l\left( \mathbf{x}\right)}\right) \right)
\end{multline}



\begin{figure}[h!]
    \centering
    \includegraphics[width=\columnwidth]{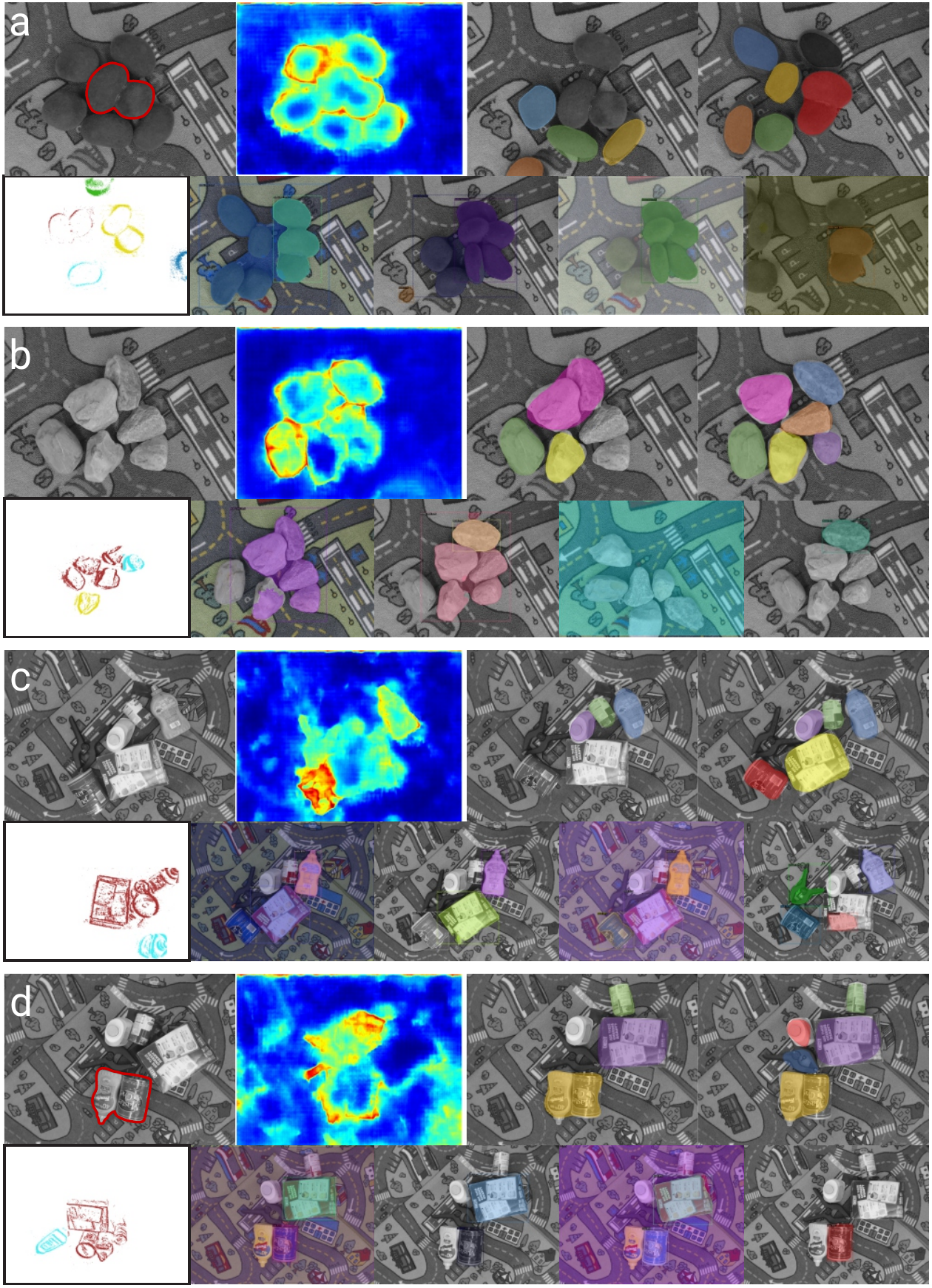}  
    \caption{For each sub-figure: First row (From left to right): Sample monochrome input image, Uncertainty in optical flow $\rho$, Segmentation hypothesis after first nudge, Final segmentation masks. Second row: (From left to right): Outputs of 0-MMS \cite{0-MMS}, PointRend (color input), PointRend (mono input), Mask-RCNN (color input), Mask-RCNN (mono input). Note that in (a) and (d), the objects highlighted with a red boundary in the top left image of the respective sequences are `glued' together and are considered to be adversarial samples. \textit{This image is viewed best in color at 400\% zoom on a computer screen.}}
    \label{fig:QualitativeEvaluation}
\end{figure}

Here, $\epsilon$ is a regularization value for avoiding numerical instability and is chosen to be $10^{-3}$. Also note that, the uncertainty is obtained at every level $l$ but we utilize the average values across levels scaled by $\alpha_l$ as an input for our framework. Refer to first row and second column of Fig. \ref{fig:QualitativeEvaluation} for the uncertainty outputs from our network. We train our model on \textit{ChairThingsMix} \cite{li2019rainflow}. \textit{We do not retrain or fine-tune the PWC-Net on any of the data used in our experiments.}




\section{Experimental Results and Discussion}
\label{sec:Results}
\subsection{Description of robot platforms -- Aerial Robot and UR10}
Our hardware setup consists of a Universal Robot UR10 and a custom-built quadrotor platform -- PRGLabrador500$\alpha$\footnote{\url{https://github.com/prgumd/PRGFlyt/wiki}} (Refer Fig. \ref{fig:Banner}). A 3D-printer end-effector is mounted on both the robots for the nudging motion. PRGLabrador500$\alpha$ is built on a S500 frame with DJI F2312 960KV motors and 9450$\times$2 propellers. The lower level attitude and position hold controller is handled by ArduCopter 4.0.6 firmware running on Holybro Kakute F7 flight controller, bridged with an optical flow sensor and a one beam LIDAR as the altimeter source. Both the robots are equipped with a Leopard Imaging M021 camera with a 3mm lens that captures the monochrome image frames at $800 \times 600$ px resolution and 30 frames/sec for our segmentation framework. All the higher level navigational commands are sent by the companion computer (NVIDIA Jetson TX2 running Linux for Tegra$^{\text{\textregistered}}$) for both the robots.

\subsection{Quantitative Evaluation}

\subsubsection{Evaluation Sequences}
\label{subsection:EvaluationSequences}
We test our \textit{NudgeSeg} framework on four sequences of objects. For evaluation of our framework, we average the results over 25 trial runs for each sequence. 

For each iteration of an evaluation sequence, $N_i$ objects are randomly chosen from a given range of objects $N$ ($N_i \le N$) and are placed on a table in a random configuration space. Now, let $\mathpzc{N}$ be the total number of nudges required to solve the segmentation task for a given configuration and $\mathpzc{N}_{\text{avg}}$ be the average number of nudges for all the trials of a sequence. Refer to Table \ref{tab:Sequences} for details on the sequences and Fig. \ref{fig:EvaluationSequences} for a visual depiction of the sequences. It is important to note that objects in sequences \texttt{GrassMoss} and \texttt{YCB-attached} (Fig. \ref{fig:EvaluationSequences} (a) and (d)) contains adversarial samples of objects that are been permanently ``glued'' together.\\ 

\subsubsection{State-of-the-art Methods}
We compare our results with three state-of-the-art methods: 0-MMS \cite{0-MMS}, Mask-RCNN \cite{he2017mask} and PointRend \cite{kirillov2020pointrend}. 

0-MMS \cite{0-MMS} utilizes an event camera rather than a classical camera for segmenting the moving objects. These event cameras output asynchronous log intensity differences caused by relative motion at microsecond latency. 
This class of sensor differs from the traditional camera as it outputs events which are sparse and have high temporal resolution. It relies on large displacement of objects between time intervals. As event data is highly dependent on motion, we induced \textit{large motion} in the objects by a single large nudge as presented in the original work.

\begin{figure}
    \centering
    \includegraphics[width=\columnwidth]{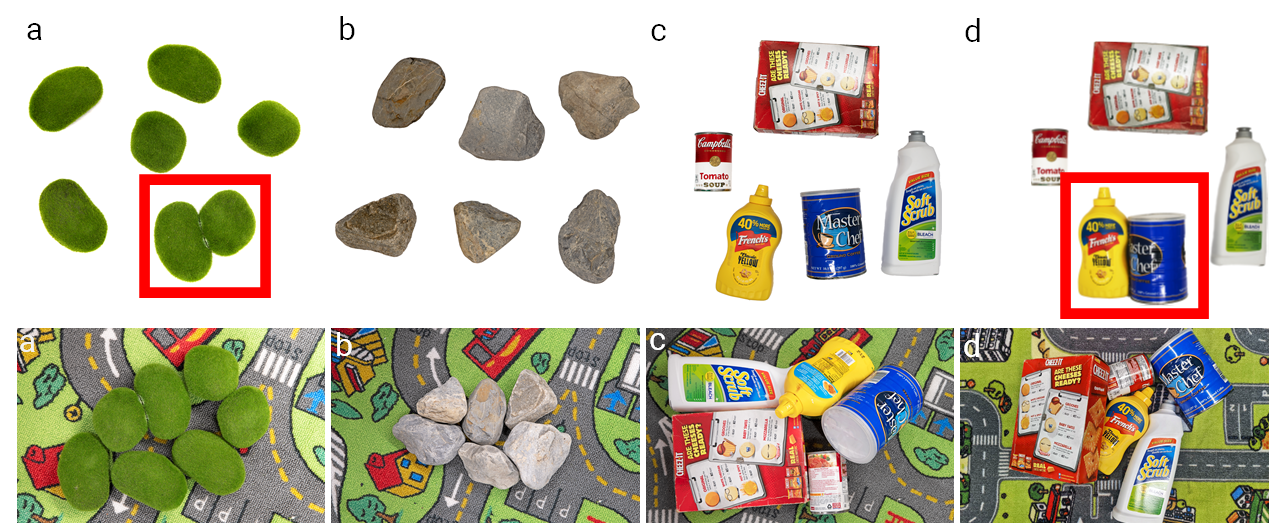}  
    \caption{Top Row: Sample objects used in Table \ref{tab:Sequences} as the evaluation sequences. Bottom Row: Sample cluttered scene for each sequence.}
    \label{fig:EvaluationSequences}
\end{figure}

We also compare our results with single frame passive segmentation methods (no induced active movement): Mask-RCNN and PointRend which were trained on the \texttt{train2017} ($\sim$123K images) of MS-COCO \cite{lin2014microsoft} dataset containing common objects. Note that, YCB object sequences \cite{calli2015ycb} as shown in Figs. \ref{fig:EvaluationSequences}{\color{red}c} and \ref{fig:EvaluationSequences}{\color{red}d} form a subset of the classes in MS-COCO dataset. However, the sequences shown in Figs. \ref{fig:EvaluationSequences}{\color{red}a} and \ref{fig:EvaluationSequences}{\color{red}b} are zero-shot samples (classes not present in MS-COCO dataset). \\


\begin{table}[t!]
\centering
\caption{Description of Evaluation Sequences.}
\resizebox{0.75\columnwidth}{!}{
\label{tab:Sequences}
\begin{tabular}{llllll}
\toprule
Sequence Name & $N$ & $\mathpzc{N}_{\text{avg}}$ & Reference Fig. \\
  \hline\\[-5pt]
  
  \texttt{GrassMoss} & 5-8 & 5.7 &  \ref{fig:EvaluationSequences}{\color{red}a}\\
  \texttt{Rocks} & 5-7 & 5.2 &  \ref{fig:EvaluationSequences}{\color{red}b}\\
  \texttt{YCB} & 5-9 & 6.3 &  \ref{fig:EvaluationSequences}{\color{red}c}\\
  \texttt{YCB-attached} & 4-8 & 4.8 &  \ref{fig:EvaluationSequences}{\color{red}d}\\
 \bottomrule
\end{tabular}}
\end{table}

\subsubsection{Evaluation Metrics}

Before we evaluate and compare the performance of \textit{NudgeSeg} with the aforementioned methods, let us first define the evaluation metrics used.
Intersection over Union (IoU) is one of the most common and standard evaluation criteria use for comparing different segmentation methods. IoU is given by:
    
\begin{equation}
    IoU = (\mathcal{D} \cap \mathcal{G}) / (\mathcal{D} \cup \mathcal{G})
\end{equation}

where $\mathcal{D}$ is the predicted mask and $\mathcal{G}$ is the ground truth mask. We define Detection Rate (DR) on the basis of IoU for each cluster as
\begin{equation}
\text{Success} := IoU \geq \tau ;\,\, \text{DR}  = \frac{\text{Num. Success}}{\text{Num. Trials}}
\end{equation}

We define DR at two accuracy levels with $\tau$ of 0.5 and 0.75 which are denoted as $\text{DR}_{50}$ and $\text{DR}_{75}$ respectively. For passive segmentation methods, we compute $IoU_i$, $\text{DR}_{50, i}$ and $\text{DR}_{75, i}$ for an image after ever nudge (indexed as $i$) along with the initial configuration. The final $IoU$, $\text{DR}_{50}$ and $\text{DR}_{75}$ for every trial (which includes multiple nudges) are given by the highest accuracy among all the time indexes for a fair comparison. Finally, we compute the average results across trials for each scenario as mentioned before.\\






\subsubsection{Observations}

Now, let us compare \textit{NudgeSeg} with other state-of-the-art segmentation methods using the aforementioned evaluation metrics. Table \ref{tab:Accuracy} shows the performance of our method on different sequences and compares it with different methods (Refer \ref{subsection:EvaluationSequences}). We compare PointRend and Mask-RCNN segmentation methods on both RGB and monochrome images. Since \textit{NudgeSeg} has more motion cues by iterative nudging the cluttered scene, it performs better than 0-MMS which relies on motion cues from only one large nudge. Furthermore, the active segmentation approaches -- \textit{NudgeSeg} framework and 0-MMS substantially outperform the passive segmentation methods -- PointRend and Mask-RCNN at the zero-shot samples which highlights the usability of active methods. This is due to the iterative segmentation propagation and the elegance of the sensor-motor loops introduced (nudge the part of the scene where we want to gather more information). Such an active method can be used to train a passive method like Mask-RCNN or PointRend in a self-supervised manner, thereby mimicking the memory in animals.

Since, the generalization performance of most passive segmentation methods are subpar to that of the active methods, however their masks are sharper during correct segmentation because they incorporate memory information. To capture this essence, we define a new metric $IoU_{s}$ which is given as the Intersection over Union for successfully detected objects. 
As we stated before, we obtain the segmentation results after every nudge and the highest accuracy results are used. Finally, we compute the average  results  across  trials  for  each  scenario  as  mentioned before.



From Table \ref{tab:Accuracy}{\color{red}(d)}, we see that the masks predicted by Mask-RCNN and PointRend are sharper (better overlap with the ground truth) when the adversarial objects are excluded and they approach  (decrease) the results of the active methods since the \textit{IoU} values are averaged for success of \textit{IoU} $\geq$ 0.5. We can clearly observe that incorporating memory information can significantly boost the segmentation performance around the edges which are useful for grasping. Also, note that results for  Mask-RCNN and PointRend are slightly lower than the active methods when no color information is given as the input, showing that these networks rely on color boundaries for segmentation. \\


\begin{table}[t!]
\centering
\caption{Evaluation with different segmentation methods for multiple sequences.}
\resizebox{\columnwidth}{!}{
\label{tab:Accuracy}
\begin{tabular}{c c c c c c c}
\toprule
Sequence & \textit{NudgeSeg} & 0-MMS \cite{0-MMS} & \multicolumn{2}{c}{PointRend \cite{kirillov2020pointrend} } & \multicolumn{2}{c}{ Mask-RCNN\cite{he2017mask}}\\
\cline{2-7}\\[-6pt]
Sensor Used & Mono & Event & Color & Mono & Color & Mono \\
\midrule
\multicolumn{7}{c}{(a) $\text{IoU}\uparrow$}\\
\midrule
GrassMoss & 0.82 & 0.77 & 0.14 & 0.14 & 0.12 & 0.10\\
Rocks & 0.87 & 0.63 & 0.16 & 0.17 & 0.11 & 0.13\\
YCB & 0.68 & 0.58 & 0.44 & 0.42 & 0.36 & 0.39\\
YCB-attached & 0.70 & 0.61 & 0.38 & 0.35  & 0.32 & 0.32\\
\midrule

\multicolumn{7}{c}{(b) $\text{DR}_{50} (\%)\uparrow$}\\
\midrule
GrassMoss & 89.6 & 64.3 & 11.1 & 9.4 & 8.2 & 6.7\\
Rocks & 94.1 & 30.2 & 14.7 & 14.1 & 9.5 & 7.7\\
YCB & 80.9 & 28.4 & 42.4 & 40.6 & 36.1 & 39.3\\
YCB-attached & 82.0 & 32.2 & 37.7 & 31.4 & 32.1 & 34.9\\
\midrule

\multicolumn{7}{c}{(c) $\text{DR}_{75} (\%)\uparrow$}\\
\midrule
GrassMoss & 86.3 & 64.4 & 10.1 & 9.9 & 7.4 & 6.3\\
Rocks & 91.1 & 30.9 & 13.5 & 13.2  & 7.2 & 6.1\\
YCB & 74.5 & 42.3 & 38.8 & 35.1  & 32.9 & 34.2\\
YCB-attached & 76.2 & 32.4 & 33.1 & 27.0 & 27.2 & 29.3\\
\midrule
  
\multicolumn{7}{c}{(d) $\text{IoU}_s\uparrow$ (for $\text{DR}_{50}$)}\\
\midrule
GrassMoss & \textbf{0.88} & 0.84 & 0.83 (\textbf{0.91}) & 0.80 (0.87) & 0.81 (0.90) & 0.79 (0.88)\\
Rocks & 0.89 & 0.80 & \textbf{0.97} & 0.95 & 0.91 & 0.88\\
YCB & 0.77 & 0.70 & \textbf{0.96} & 0.88 & 0.92 & 0.85\\
YCB-attached & 0.75 & 0.72 & \textbf{0.79} & 0.77 & 0.75 & 0.72\\
\bottomrule
\end{tabular}}
\begin{tiny}
Note: $(\cdot)$ represents the $\text{IoU}_s$ after the removal of adversarial examples in \texttt{GrassMoss} sequence.
\end{tiny}
\end{table}

\begin{figure*}[h!]
    \centering
    \includegraphics[width=\textwidth]{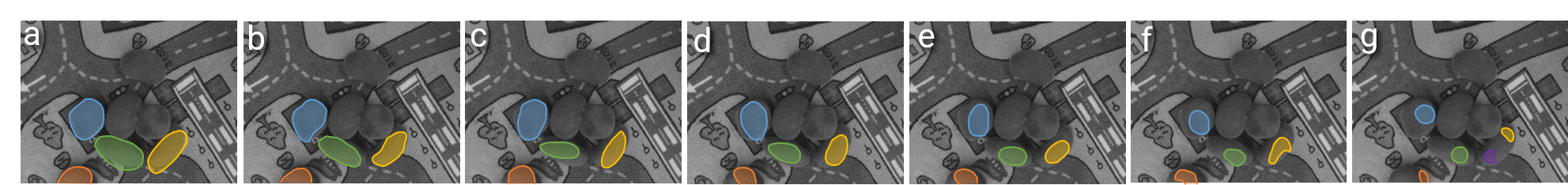}  
    \caption{Qualitative Results with (a) no error, $\epsilon_{\mathcal{A}}=0$, $\epsilon_{\mathcal{M}}=$ (b) $\pm$5\%, (c) $\pm$10\%, (d) $\pm$20\%, $\epsilon_{\mathcal{M}}=0$, $\epsilon_{\mathcal{A}}=$ (e) $\pm10^\circ$, (f) $\pm20^\circ$, (g) $\pm30^\circ$.}
    \label{fig:FlowError}
\end{figure*}

\subsubsection{Analysis}
For most robotics applications, the performance of the algorithm is also governed by the camera sensor quality. Missing data, false colors, bad contrast, motion blur and low image resolution generally cause robotics algorithms to fail. Such erroneous sensor behaviour often leads to a large amount of errors in fundamental properties like optical flow. To test the limitations of our framework, we artificially inject error in optical flow, both in angle and magnitude separately; and evaluate our framework performance under these conditions. We add uniform noise $\mathcal{U}\left(\epsilon\right)$ at each pixel to both $\mathcal{A}$ and $\mathcal{M}$ (Refer \ref{subsection:InteractiveSeg} for definitions) independently where $[-\epsilon,\epsilon]$ is the bound on noise. The perturbed (noise induced) magnitude and angle of the flow vectors are denoted as $\mathcal{\widetilde{M}}$ and $\mathcal{\widetilde{A}}$ respectively and are given by

\begin{equation}
\mathcal{\widetilde{M}}_{\mathbf{x}} = \left(1 + \frac{\mathcal{U}(\epsilon_{\mathcal{M}})}{100}\right) \mathcal{M}_{\mathbf{x}}
\end{equation}
\begin{equation}
\mathcal{\widetilde{A}}_{\mathbf{x}} = \left(1 + \mathcal{U}(\epsilon_{\mathcal{A}})\right) \mathcal{A}_{\mathbf{x}}
\end{equation}

\begin{table}[t!]
\centering
\caption{Evaluation of \texttt{GrassMoss} sequence with different amount of errors in $\mathcal{A}$ and $\mathcal{M}$.}
\resizebox{\columnwidth}{!}{
\label{tab:FlowError}
\begin{tabular}{llllllll}
\toprule
Err. Metric & No Error & $\epsilon_M=5$ & $\epsilon_M=10$ & $\epsilon_M=20$ & $\epsilon_A=10^\circ$ & $\epsilon_A=20^\circ$ & $\epsilon_A=30^\circ$\\
\midrule
$IoU \uparrow$  & 0.82 & 0.77 & 0.70 & 0.64 & 0.62 & 0.41 & 0.23\\
$\text{DR}_{50} (\%) \uparrow$ & 89 & 82 & 74 & 68 & 61 & 60 & 49 \\
$\text{DR}_{75} (\%) \uparrow$ & 86 & 77 & 69 & 60 & 46 & 37 & 17 \\
\bottomrule
\end{tabular}}\\[1pt]
\begin{tiny}
If the injected error is not stated explicitly, it is taken to be zero. 
\end{tiny}
\end{table}

We evaluate our framework on a different $\epsilon_{\mathcal{M}}$ and $\epsilon_{\mathcal{A}}$ values of $\{0,5,10,20\}\%$ and $\{0,10,20,30\}^\circ$ respectively. Fig. \ref{fig:FlowError} shows a qualitative result on how  error in optical flow affects the first segmentation hypothesis $\mathcal{H}_1$. Note that the error in flow is added to the PWC-Net output which has the angle error of $\sim$5\% as compared to the ground truth optical flow. The $IoU$ performance in $\mathcal{H}_1$ significantly decreases when the noise is added to $\mathcal{A}$ as opposed to $\mathcal{M}$. This shows that optical flow angle is more important for active segmentation methods as compared to magnitude and such a distinction in evaluation is often missing in most optical flow works. 

Table \ref{tab:FlowError} shows the performance of \texttt{GrassMoss} sequence with different errors in $\mathcal{A}$ and $\mathcal{M}$. The $DR_{75}$ drops significantly from $86\%$ to $\sim$17\% with a $\pm30^\circ$ error in $\mathcal{A}$ (about $6\times$ the error than in PWC-Net). Clearly, accurate optical flow computation is essential for active and interactive segmentation approaches. Hence, speeding up the neural networks by quantizing or reducing the number of parameters will dictate the performance of interactive segmentation methods.


We also evaluate passive segmentation methods on the sample non-cluttered images shown in Fig. \ref{fig:EvaluationSequences} (top row). Both PointRend and Mask-RCNN have similar performance in \texttt{GrassMoss} and \texttt{Rocks} sequences but perform substantially better for \texttt{YCB} resulting in about $94\%$ $\text{DR}_{50}$ for both the methods showing that the background clutter and the object occlusions can significantly affect the segmentation performance.

\section{Discussion and Future Directions}
\label{sec:Discussion}
Depending on the Size, Weight, Area and Power (SWAP) constraints, the robot may be equipped with multiple sensors including a depth camera. Such sensors would be an essential element to allow nudging in three dimensions. In such a case, \textit{where to nudge?} would also be computed using a path planner to avoid collisions. 

However, one can model the \textit{where to nudge?} problem using reinforcement learning to obtain a more elegant solution. In particular, we envision a reward function based on the success criterion of the \textit{NudgeSeg} framework for delayed rewards. This can help learn more generalized optical flow along with segmentation which would work better for zero-shot objects in a true robotics setting by utilizing the sensori-motor loops. Here, the reinforcement agent would predict where to nudge and it's estimated reward in a manner that is agnostic to the actual class of the objects.


To improve our framework further, one can utilize a soft suction gripper which can adapt to different morphologies of the object shape. In particular, the suction gripper can be used after each nudging step to pick the unknown object and ``remove'' it from the cluttered pile to make the perception problem even easier.

By combining the two aforementioned directions, we believe that the active--interactive segmentation model is the future which aligns beautifully with the life-long learning paradigm \cite{parisi2019continual}.

\section{Conclusions}
\label{sec:Conclusions}

We present an active-interactive philosophy for segmenting unknown objects from a cluttered scene by repeatedly `nudging' the objects and moving them to obtain additional motion cues. These motion cues (in the form of optical flow) are used to find and refine segmentation hypothesis at every step. Our approach only uses a monochrome monocular camera and performs better than the current state-of-the-art object segmentation methods by a large margin for zero shot samples. We successfully demonstrate and test our approach to segment novel objects in various cluttered scenes and provide an extensive comparison with passive and motion segmentation methods on different mobile robots: a quadrotor and a robotic arm. We show an impressive average detection rate of over $86\%$ on zero-shot samples. We firmly believe that such a method can serve as the first step to learn novel objects to enable a true lifelong learning system.

\section*{Acknowledgement}
The support of the National Science Foundation under grants BCS 1824198 and OISE  2020624 and the support of  Office of Naval Research under grant award N00014-17-1-2622 are gratefully acknowledged. We would like to also thank Samsung Electronics for the event cameras used in this experiment.
\bibliographystyle{unsrt}
\bibliography{Ref}

\begin{thebibliography}{10}

\bibitem{egelhaaf2012spatial}
Martin Egelhaaf, Norbert Boeddeker, Roland Kern, Rafael Kurtz, and Jens~Peter
  Lindemann.
\newblock Spatial vision in insects is facilitated by shaping the dynamics of
  visual input through behavioral action.
\newblock {\em Frontiers in neural circuits}, 6:108, 2012.

\bibitem{long2015fully}
Jonathan Long, Evan Shelhamer, and Trevor Darrell.
\newblock Fully convolutional networks for semantic segmentation.
\newblock In {\em Proceedings of the IEEE conference on computer vision and
  pattern recognition}, pages 3431--3440, 2015.

\bibitem{he2017mask}
Kaiming He, Georgia Gkioxari, Piotr Doll{\'a}r, and Ross Girshick.
\newblock Mask r-cnn.
\newblock In {\em Proceedings of the IEEE international conference on computer
  vision}, pages 2961--2969, 2017.

\bibitem{chen2019tensormask}
Xinlei Chen, Ross Girshick, Kaiming He, and Piotr Doll{\'a}r.
\newblock Tensormask: A foundation for dense object segmentation.
\newblock In {\em Proceedings of the IEEE/CVF International Conference on
  Computer Vision}, pages 2061--2069, 2019.

\bibitem{kirillov2020pointrend}
Alexander Kirillov, Yuxin Wu, Kaiming He, and Ross Girshick.
\newblock Pointrend: Image segmentation as rendering.
\newblock In {\em Proceedings of the IEEE/CVF conference on computer vision and
  pattern recognition}, pages 9799--9808, 2020.

\bibitem{torr1998geometric}
Philip~HS Torr.
\newblock Geometric motion segmentation and model selection.
\newblock {\em Philosophical Transactions of the Royal Society of London.
  Series A: Mathematical, Physical and Engineering Sciences},
  356(1740):1321--1340, 1998.

\bibitem{tron2007benchmark}
Roberto Tron and Ren{\'e} Vidal.
\newblock A benchmark for the comparison of 3-d motion segmentation algorithms.
\newblock In {\em 2007 IEEE conference on computer vision and pattern
  recognition}, pages 1--8. IEEE, 2007.

\bibitem{517092}
Y.~{Weiss} and E.~H. {Adelson}.
\newblock A unified mixture framework for motion segmentation: incorporating
  spatial coherence and estimating the number of models.
\newblock In {\em Proceedings CVPR IEEE Computer Society Conference on Computer
  Vision and Pattern Recognition}, pages 321--326, 1996.

\bibitem{jepson1993mixture}
Allan Jepson and Michael~J Black.
\newblock Mixture models for optical flow computation.
\newblock In {\em Proceedings of IEEE Conference on Computer Vision and Pattern
  Recognition}, pages 760--761. IEEE, 1993.

\bibitem{bideau2018best}
Pia Bideau, Aruni RoyChowdhury, Rakesh~R Menon, and Erik Learned-Miller.
\newblock The best of both worlds: Combining cnns and geometric constraints for
  hierarchical motion segmentation.
\newblock In {\em Proceedings of the IEEE Conference on Computer Vision and
  Pattern Recognition}, pages 508--517, 2018.

\bibitem{bideau2016s}
Pia Bideau and Erik Learned-Miller.
\newblock It’s moving! a probabilistic model for causal motion segmentation
  in moving camera videos.
\newblock In {\em European Conference on Computer Vision}, pages 433--449.
  Springer, 2016.

\bibitem{wulff2017optical}
Jonas Wulff, Laura Sevilla-Lara, and Michael~J Black.
\newblock Optical flow in mostly rigid scenes.
\newblock In {\em Proceedings of the IEEE Conference on Computer Vision and
  Pattern Recognition}, pages 4671--4680, 2017.

\bibitem{stoffregen2019event}
T.~Stoffregen et~al.
\newblock Event-based motion segmentation by motion compensation.
\newblock In {\em International Conference on Computer Vision (ICCV)}, 2019.

\bibitem{0-MMS}
Chethan~M Parameshwara, Nitin~J Sanket, Chahat~Deep Singh, Cornelia
  Ferm{\"u}ller, and Yiannis Aloimonos.
\newblock 0-mms: Zero-shot multi-motion segmentation with a monocular event
  camera.

\bibitem{ActiveVision}
J.~Aloimonos et~al.
\newblock Active vision.
\newblock {\em International journal of computer vision}, 1(4):333--356, 1988.

\bibitem{BajcsyActive}
R.~Bajcsy et~al.
\newblock Revisiting active perception.
\newblock {\em Autonomous Robots}, pages 1--20, 2017.

\bibitem{8007233}
J.~{Bohg}, K.~{Hausman}, B.~{Sankaran}, O.~{Brock}, D.~{Kragic}, S.~{Schaal},
  and G.~S. {Sukhatme}.
\newblock Interactive perception: Leveraging action in perception and
  perception in action.
\newblock {\em IEEE Transactions on Robotics}, 33(6):1273--1291, 2017.

\bibitem{katz2013clearing}
Dov Katz, Moslem Kazemi, J~Andrew Bagnell, and Anthony Stentz.
\newblock Clearing a pile of unknown objects using interactive perception.
\newblock In {\em 2013 IEEE International Conference on Robotics and
  Automation}, pages 154--161. IEEE, 2013.

\bibitem{hausman2012segmentation}
Karol Hausman, Christian Bersch, Dejan Pangercic, Sarah Osentoski, Zoltan-Csaba
  Marton, and Michael Beetz.
\newblock Segmentation of cluttered scenes through interactive perception.
\newblock In {\em ICRA Workshop on Semantic Perception and Mapping for
  Knowledge-enabled Service Robotics, St. Paul, MN, USA}, 2012.

\bibitem{agrawal2016learning}
Pulkit Agrawal, Ashvin Nair, Pieter Abbeel, Jitendra Malik, and Sergey Levine.
\newblock Learning to poke by poking: Experiential learning of intuitive
  physics.
\newblock {\em arXiv preprint arXiv:1606.07419}, 2016.

\bibitem{TamimAsfour}
D.~{Schiebener}, J.~{Schill}, and T.~{Asfour}.
\newblock Discovery, segmentation and reactive grasping of unknown objects.
\newblock In {\em 2012 12th IEEE-RAS International Conference on Humanoid
  Robots (Humanoids 2012)}, pages 71--77, 2012.

\bibitem{natale2004learning}
Lorenzo Natale, Giorgio Metta, and Giulio Sandini.
\newblock Learning haptic representation of objects.
\newblock In {\em International Conference on Intelligent Manipulation and
  Grasping}, page~43, 2004.

\bibitem{pathak2018learning}
Deepak Pathak, Yide Shentu, Dian Chen, Pulkit Agrawal, Trevor Darrell, Sergey
  Levine, and Jitendra Malik.
\newblock Learning instance segmentation by interaction.
\newblock In {\em Proceedings of the IEEE Conference on Computer Vision and
  Pattern Recognition Workshops}, pages 2042--2045, 2018.

\bibitem{horn1981determining}
Berthold~KP Horn and Brian~G Schunck.
\newblock Determining optical flow.
\newblock {\em Artificial intelligence}, 17(1-3):185--203, 1981.

\bibitem{gast2018lightweight}
Jochen Gast and Stefan Roth.
\newblock Lightweight probabilistic deep networks.
\newblock In {\em Proceedings of the IEEE Conference on Computer Vision and
  Pattern Recognition}, pages 3369--3378, 2018.

\bibitem{ester1996density}
Martin Ester, Hans-Peter Kriegel, J{\"o}rg Sander, Xiaowei Xu, et~al.
\newblock A density-based algorithm for discovering clusters in large spatial
  databases with noise.
\newblock In {\em Kdd}, volume~96, pages 226--231, 1996.

\bibitem{sun2018pwc}
Deqing Sun, Xiaodong Yang, Ming-Yu Liu, and Jan Kautz.
\newblock Pwc-net: Cnns for optical flow using pyramid, warping, and cost
  volume.
\newblock In {\em Proceedings of the IEEE conference on computer vision and
  pattern recognition}, pages 8934--8943, 2018.

\bibitem{li2019rainflow}
Ruoteng Li, Robby~T Tan, Loong-Fah Cheong, Angelica~I Aviles-Rivero, Qingnan
  Fan, and Carola-Bibiane Schonlieb.
\newblock Rainflow: Optical flow under rain streaks and rain veiling effect.
\newblock In {\em Proceedings of the IEEE/CVF International Conference on
  Computer Vision}, pages 7304--7313, 2019.

\bibitem{lin2014microsoft}
Tsung-Yi Lin, Michael Maire, Serge Belongie, James Hays, Pietro Perona, Deva
  Ramanan, Piotr Doll{\'a}r, and C~Lawrence Zitnick.
\newblock Microsoft coco: Common objects in context.
\newblock In {\em European conference on computer vision}, pages 740--755.
  Springer, 2014.

\bibitem{calli2015ycb}
Berk Calli, Arjun Singh, Aaron Walsman, Siddhartha Srinivasa, Pieter Abbeel,
  and Aaron~M Dollar.
\newblock The ycb object and model set: Towards common benchmarks for
  manipulation research.
\newblock In {\em 2015 international conference on advanced robotics (ICAR)},
  pages 510--517. IEEE, 2015.

\bibitem{parisi2019continual}
German~I Parisi, Ronald Kemker, Jose~L Part, Christopher Kanan, and Stefan
  Wermter.
\newblock Continual lifelong learning with neural networks: A review.
\newblock {\em Neural Networks}, 113:54--71, 2019.

\end{thebibliography}
\end{document}